\documentclass[journal]{IEEEtran}
\usepackage[utf8]{inputenc}
\usepackage[english]{babel}
\usepackage{color}
\usepackage[table]{xcolor}
\usepackage{multirow}
\usepackage{float}
\usepackage{graphics} 
\usepackage{graphicx}
\usepackage{caption}
\usepackage{multicol}
\usepackage{subcaption}
\usepackage{color, colortbl}
\usepackage[]{algorithm2e}
\usepackage{amsmath} 
\usepackage{array}

\begin{document}

\title{Automated Evaluation of Semantic Segmentation Robustness for Autonomous Driving}

\author{Wei Zhou,~\IEEEmembership{Member,~IEEE,}
        Julie Stephany Berrio,~\IEEEmembership{Member,~IEEE,} \\
        Stewart Worrall,~\IEEEmembership{Member,~IEEE,}
        and Eduardo Nebot,~\IEEEmembership{Member,~IEEE}
        
\thanks{W. Zhou, J. Berrio, S. Worrall, and E. Nebot are with the 
Australian Centre for Field Robotics (ACFR) at the University of Sydney, NSW, Australia. E-mails: {\tt \{w.zhou, j.berrio, s.worrall, e.nebot\}@acfr.usyd.edu.au}.}
\thanks{Manuscript received on October XX, 2018; revised on Month XX, 2018.}}

\maketitle

\begin{abstract}

One of the fundamental challenges in the design of perception systems for autonomous vehicles is validating the performance of each algorithm under a comprehensive variety of operating conditions. In the case of vision-based semantic segmentation, there are known issues when encountering new scenarios that are sufficiently different to the training data. In addition, even small variations in environmental conditions such as illumination and precipitation can affect the classification performance of the segmentation model. Given the reliance on visual information, these effects often translate into poor semantic pixel classification which can potentially lead to catastrophic consequences when driving autonomously. This paper presents a novel method for analysing the robustness of semantic segmentation models and provides a number of metrics to evaluate the classification performance over a variety of environmental conditions. The process incorporates an additional sensor (lidar) to automate the process, eliminating the need for labour-intensive hand labelling of validation data. The system integrity can be monitored as the performance of the vision sensors are validated against a different sensor modality. This is necessary for detecting failures that are inherent to vision technology. Experimental results are presented based on multiple datasets collected at different times of the year with different environmental conditions. These results show that the semantic segmentation performance varies depending on the weather, camera parameters, existence of shadows, etc.. The results also demonstrate how the metrics can be used to compare and validate the performance after making improvements to a model, and compare the performance of different networks.

\end{abstract}


%
\IEEEpeerreviewmaketitle

\section{Introduction}

\IEEEPARstart{A}{utonomous} vehicles require measurably reliable and comprehensive information of the surroundings in order to make safe driving decisions. Semantic segmentation is a type of deep neural network model that assigns class labels to every pixel in the camera image, providing a high-level understanding of the scene using a similar representation to a human observer. There is a significant amount of prior work using this type of model that achieves either high accuracy~\cite{deeplabv2,pspnet,deeplabv3} when measured against a public benchmark~\cite{cityscapes}, or high computational efficiency~\cite{enet,erfnet, bonnet} to enable real-time implementation. 

\begin{figure}[t]
\centering
\begin{subfigure}{.95\columnwidth}
  \centering
  \includegraphics[width=.97\linewidth]{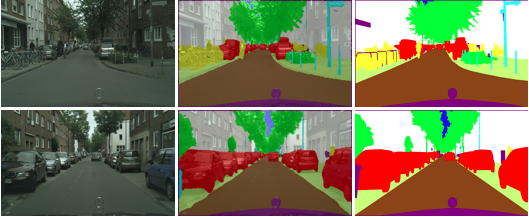}
  \caption{\small Sample semantic segmentation on Cityscapes \cite{cityscapes} validation set. From left to right: original images, predicted results from ENet model~\cite{itsc_work}, ground truth labels.}
  \label{fig:cs seg}
\end{subfigure}
\begin{subfigure}{.95\columnwidth}
  \centering
  \includegraphics[width=.97\linewidth]{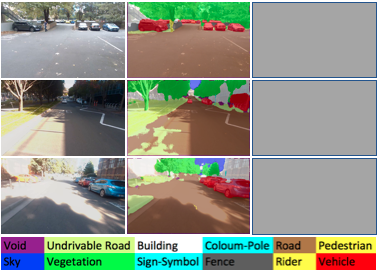}
  \caption{\small Semantic segmentation results on the University of Sydney (USYD) campus with different shadows, illumination, camera conditions, etc. From left to right: original images, predicted results from ENet model~\cite{itsc_work}, no ground truth labels.}
  \label{fig:usyd seg}
\end{subfigure}
\caption{\small Sample semantic segmentation images from different datasets. }
\label{fig:questions}
\end{figure}

Consistent with a common concern of deep neural networks, semantic segmentation models trained on publicly labeled datasets such as~\cite{cityscapes,kitti,mapillary} do not usually adapt well to new environments due to variations in scene structure, sensor configuration, weather conditions, etc. when compared to the training datasets. Further fine-tuning of a model using locally labelled data is commonly required to improve the model generalisation for various conditions. 
This process is challenging as labelling data is extremely labour intensive, and the process of selecting representative images requires significant expertise. In addition, it is critically important to validate the performance of a model across a range of locations and in a variety of weather/illumination conditions. This drastically increases the number of labelled images required to reliably validate the semantic segmentation model.  

Fig.~\ref{fig:cs seg} shows both the output of a semantic model and the original ground truth labels for a publicly available dataset. The comparison of the model output with the ground truth labels is commonly used to validate the performance. Fig.~\ref{fig:usyd seg} shows samples from our local operating environment under various environmental conditions. It is not viable to manually annotate sufficient images with ground truth labels to cover the full range of possible camera exposure settings and varying environmental conditions. This would require an order of magnitude increase in validation samples to cover different conditions across each time of day, and time of year.
To perform this validation in a comprehensive manner, it is necessary to examine new methods to make this a tractable problem. To the best of the authors' knowledge, this topic has barely attracted any attention in the research community due to the complexity and cost of generating ground truth labels.

In this paper, we propose to address the validation problem by incorporating an alternative sensing modality (lidar) to automatically generate ground truth labels for evaluation. 
As the estimation of the drivable road area is one of the most essential classes for autonomous vehicles, a lidar-based road extraction approach is proposed to validate specifically the `Road' class of a semantic segmentation model. 
This complementary sensing modality is not as prone to varying lighting or shading conditions, allowing the validation of the model performance across a range of possible scenarios.

To evaluate the real-world performance of a semantic model, it is necessary to collect data of the proposed area of operation with varied time of day throughout the year.
For our research, the target area is the surroundings of the University of Sydney campus (USYD). We have collected datasets covering the entire campus periodically over an extended period that has captured a wide range of conditions.
This data is a valuable and unique resource to address the fundamental reliability issue of semantic segmentation for autonomous vehicles.
In this paper we present multiple validation methods to demonstrate the performance and failure cases of several semantic models when operating across different environmental conditions, and across a variety of locations.

The major contributions in this paper are:
\begin{itemize}

  \item A semantic segmentation validation pipeline to automatically validate any semantic model across different environments without additional hand labelling of data. The proposed method provides new metrics to perform an analysis of the classification failure rate, and to determine the quality of the model using multiple datasets. A local dataset containing a number of drives throughout the year with varied conditions is under preparation and will be released soon.
  \item Automatic generation of ground-truth labelled data for the `Road' class using lidar as a complementary sensing modality. This sensor is immune to many of the illumination conditions such as strong shadows that are one of the main sources of error in vision-based semantic segmentation. The proposed method completely avoids time-consuming hand labelling. 
  \item Several applications utilising the proposed pipeline are presented to show the significance of the validation process. Normally, different network architectures are compared against publicly available datasets with a limited range of camera and environmental conditions. In this paper, we present the results of the semantic classification when operating under a variety of environment conditions that are commonly experienced in real-world scenarios. This clearly demonstrates that it is insufficient to measure the performance of a model by only testing against public datasets. The presented method can also be used to validate deep learning augmentation techniques applied during training. The performance metrics can further be associated with global position information that enables a location-based comparison. This allows us to understand and compare the performance given the spatial context when validating a model.
\end{itemize}

The paper is organized as follows: Section \ref{sec:related work} presents background work related to vision and lidar based segmentation, and the estimation of reliability in the semantic segmentation process. Section \ref{sec:Semantic Seg Val} describes the proposed validation pipeline with a detailed description of the main components. Section \ref{sec:experiment} shows the pipeline implementation details. Section \ref{sec:Experimental Results} presents the experimental results for the validation of semantic segmentation models across varied environmental conditions. Finally, the conclusion and future work are presented in Section \ref{sec:Conclusions}.


\section{Related Work}
\label{sec:related work}

In this section, we will present an overview of the existing work that relates to the main contributions in this paper. 

\subsection{Semantic Segmentation}
Semantic segmentation is the process of allocating a semantic class label to each pixel in an image. Originating from FCN~\cite{fcn} and SegNet~\cite{segnet}, today there is significant literature and a range of models and techniques available in this area. 
DeepLabv2~\cite{deeplabv2} proposed an Atrous Spatial Pyramid Pooling (ASPP) method which segments images at different scales and organises modules either in cascade or in parallel. 
PSPNet~\cite{pspnet} also presented novel work to take local and global features into account for predictions. The pyramid pooling module in their architecture separated feature maps from ResNet~\cite{resnet} into sub-regions at different scales and obtained pooled representations for those sub-regions. By adding an auxiliary loss, they optimised the learning process and achieved state-of-the-art performance across many benchmarks. 
DeepLabv3~\cite{deeplabv3} re-evaluated the atrous convolution and proved that by varying the rate for a filter's field of view (FOV), it is possible to control the feature map resolution and obtain more information from the image. DeepLabv3+~\cite{deeplabv3+} further added a decoder to extend DeepLabv3 to refine object boundaries. 

Apart from these high-accuracy networks, there are also some architectures designed primarily for improved real-time performance. ENet~\cite{enet} down-samples images at the early stage and utilises a small decoder to achieve fast inference speed. ERFNet~\cite{erfnet} replaces the bottleneck module in ENet and proposed the use of convolutions with 1D kernels. This resulted in a significant reduction of computational cost while maintaining similar accuracy to using 2D convolutions.   

\subsection{Road Detection and Segmentation}

\subsubsection{Lidar-based algorithms}
Since lidar does not provide texture or colour information, estimating the extent of the road surface with lidar is usually achieved by extracting the geometric properties of curbs, barriers, and road boundary lines from the sensor point cloud. 

Kang et al. \cite{kang2012lidar} developed an interacting multiple model method to detect road curbs using lidar measurements. The geometric features of curb points were detected by a Hough transform-based method. Then, two Kalman filters were applied based on different hypotheses to track the existence of a road curb.  
A downward-looking lidar was used in \cite{han2012enhanced} to detect the road boundary and low obstacles. By assuming a flat ground, they extracted the line segments in polar coordinates with roll and pitch angles, and tracked road boundaries using an integrated probabilistic data association filter. 
In \cite{mark_yourself}, Bruls et al. proposed a road surface extraction method based on analyzing the normal of every lidar point within a radius of 0.35m range. The surface normal was calculated by principal component analysis and the road boundary was found when the normal is no longer perpendicular to the ground surface. 

\subsubsection{Camera-based algorithms}
As lidar points are generally sparse, camera images are widely used to estimate the road surface given their rich colour and high level contextual information. 

Extended StixelWorld~\cite{stixel} used colour information to learn models for road and obstacles. These models can be used together with disparity information to obtain vertical stacks and a planar ground representation which distinguishes road from obstacles. 
Deep neural network (DNN) models have recently demonstrated remarkable performance improvements that have been shown to outperform most traditional methods. A number of contributions in the area of DNN road segmentation are presented in \cite{dnn1,dnn2,dnn3,road_seg1,road_seg2}. 
However, their performance has not been carefully validated for real-world applications due to the lack of publicly available datasets. Also vision-based methods have been demonstrated to have poor performance under extreme illumination conditions such as shadows, direct sunlight or over/under exposed images. This variability of conditions are not fully considered in the publicly available datasets, which makes testing and validation possible for only a limited range of scenarios.

\subsubsection{Multiple sensor modalities}
Another approach to add robustness to the detection process is by combining different sensor modalities. Conditional random field (CRF) based algorithms are popular and widely applied in many road detection algorithms~\cite{cfr_multi_sensor,fusion1}.
In \cite{fusion1}, it was proposed to project the lidar point clouds into the camera image to get a height encoded image. By using a joint bilateral filter, they obtained a high-resolution height image and classified pixels into road and non-road classes accordingly.

\subsection{Segmentation Reliability}
As mentioned in \cite{camera-config}, vision-based algorithms may have very different performance with different camera configurations and environmental conditions. This aspect of reliability and integrity is of significant importance for autonomous vehicles. The application of any autonomous technology will require a verifiable processes to validate and identify failure modes for any critical algorithms.
Kim el al.~\cite{fail2} introduced a robust vehicle detection and tracking system which could work in different lighting and road conditions. They validated the detected vehicle candidates by extracting vertical edges at left and right side of the vehicle to avoid false positive errors induced by shadow patterns. 
Ramanagopal et al.~\cite{fail1} proposed an automated verification process which compares the detection performance of a pair of images taken from stereo cameras to detect errors. This process does not require labelled data, and is able to correlate the error with geospatial information. The pair of images are run through the same detector, and due to the differences in camera pose and sensitivity of the detector there are times when the resulting detector output differs. The drawback of this approach is the use of a single sensor modality in which each information source can be affected by the same common failure mode such as shadows and lighting effects. Another issue is that it is measuring the performance of the detector against itself, which is useful for making the detector less sensitive but is unable to validate the performance against unseen data.

\begin{figure}[!t]
\centerline{
\includegraphics[height=7.5cm,width=0.95\columnwidth]{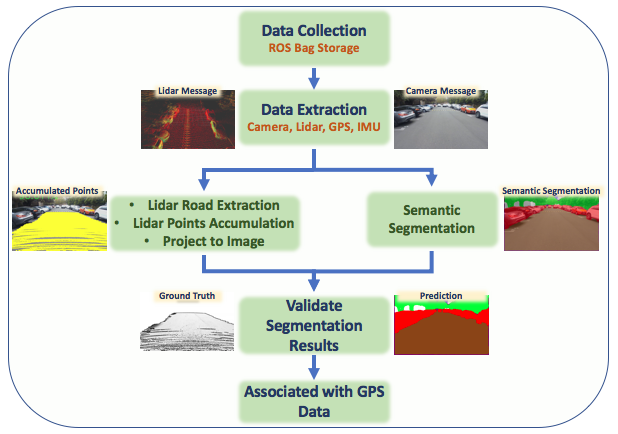}
}
\caption{\small Pipeline of semantic validation by using multiple sensor modalities. }
\label{fig:pipeline}
\end{figure}

\begin{figure*}[t]
\centering
\begin{subfigure}{.85\columnwidth}
  \centering
  \includegraphics[height=6.3cm, width=.99\linewidth]{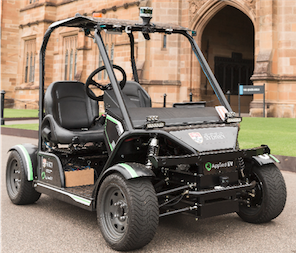}
  \caption{\small Data collection platform}
  \label{fig:car}
\end{subfigure}
\begin{subfigure}{.85\columnwidth}
  \centering
  \includegraphics[height=6.3cm, width=.99\linewidth]{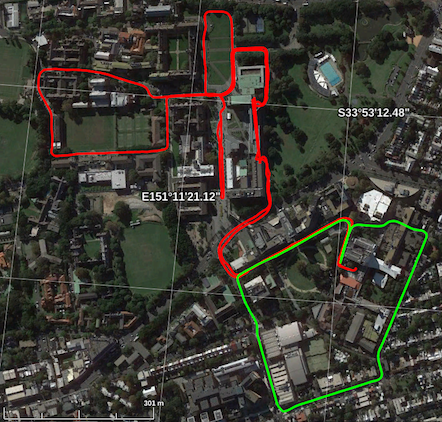}
  \caption{\small Driving trajectory}
  \label{fig:trajectory}
\end{subfigure}
\caption{\small Data collection. (a) Electrical vehicle (EV) designed to provide `last-mile' mobility around the USYD campus. (b) Data collection trajectories. Green line is the trajectory we used to collect and annotate USYD Dataset. Red line is the trajectory we repeatedly collect data every week for testing purpose.}
\label{fig:car_config}
\end{figure*}

\begin{figure*}[!t]
    \centering
    \begin{subfigure}[b]{1.1\columnwidth}
        \includegraphics[height=4.5cm, width=\columnwidth]{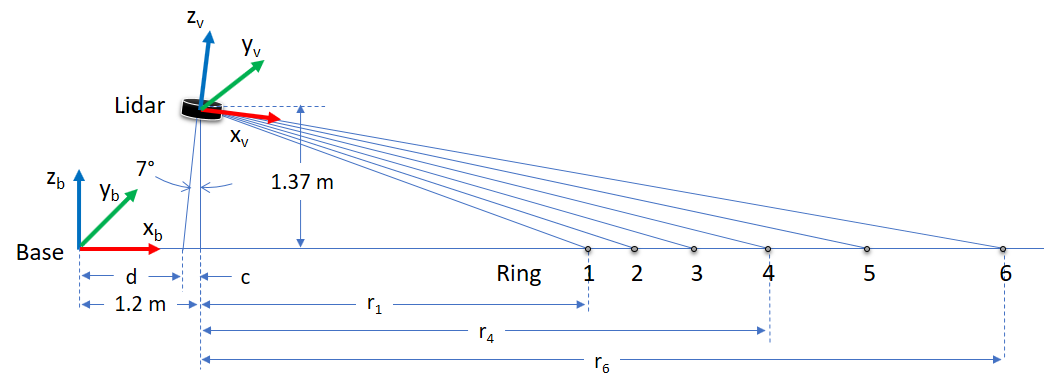}
        \caption{\small Six lidar beams used for road extraction}
        \label{fig:beam}
    \end{subfigure}
    ~
    \begin{subfigure}[b]{0.7\columnwidth}
        \includegraphics[height=4.5cm, width=\columnwidth]{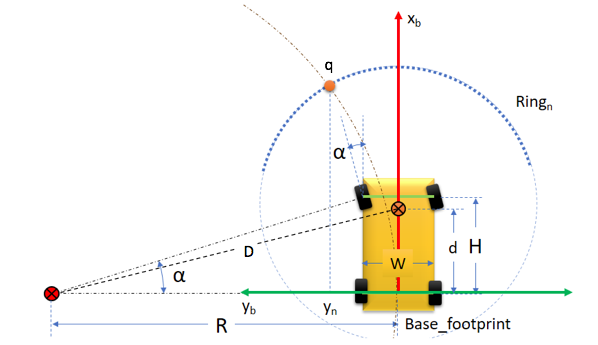}
        \caption{\small Top-down view of the vehicle and lidar ring}
        \label{fig:circle}
    \end{subfigure}

    \caption{\small Lidar road extraction.}
    \label{fig:lidar_road_algorithm}
\end{figure*}

\section{Semantic Segmentation Validation Framework}
\label{sec:Semantic Seg Val}
This section introduces the semantic validation pipeline and describes the essential components required for the implementation. 

The overall validation pipeline is shown in Fig. \ref{fig:pipeline}. The data collection platform (Fig.~\ref{fig:car}) incorporates cameras, lidar, global positioning system (GPS), and dead-reckoning information based on encoders and an inertial measurement unit (IMU). The platform is used to collect data across a range of different environmental conditions over an extended period of time. From the lidar data, we then extract the drivable road surface to use as a ground truth label and compare this label to the vision-based semantic segmentation `Road' class. In the final stage, we generate a set of metrics to evaluate the performance of the semantic segmentation against the lidar extracted road label, and associate the results with geographical information. The following sections present a detailed description of each of these components.

\subsection{Data Collection and Preparation}
\subsubsection{Vehicle setup}
To test the algorithm under different environmental conditions, we collected a weekly dataset over a 6 month period around the USYD campus using an electrical vehicle (EV) as shown in Fig.~\ref{fig:car}.
Our data collection vehicle is designed to operate using the Robot Operating System (ROS), and all collected data is stored as standard messages in ROS bags. The transforms between different coordinate systems are published as tf messages~\cite{tf} which is utilised to maintain the relationship between coordinates in a tree structure. The vehicle is equipped with the following sensing and processing capabilities:
\begin{itemize}
  \item 6 $\times$ SF3322 NVIDIA 2MP GMSL cameras, each with $100^\circ$ horizontal field of view (FOV) and $60^\circ$ vertical FOV, 1928$\times$1208$\times$3, 30Hz
  \item 1 $\times$ Velodyne VLP-16 rotating 3D laser scanner, 10Hz, mounted $7^\circ$ facing downwards
  \item Vectornav VN-100 IMU, 100Hz
  \item Wheel Encoders
  \item UBlox GPS, 1Hz
  \item NVIDIA DRIVE PX 2 computing platform
\end{itemize}
The front facing camera is tilted $15^\circ$ downwards so that the lowest pixels in the image correspond to a point 1.4 meters in front of the vehicle on flat ground, which is important to detect close obstacles. This camera is used throughout this paper to generate the semantic segmentation results and for the projection of the lidar points. The NVIDIA DRIVE PX 2 provides the computing capability to implement real-time semantic segmentation on board the vehicle.

\subsubsection{Data collection route}
In Fig.~\ref{fig:trajectory}, the red trajectory is the route traversed each week during the data collection period. The various datasets include a diverse range of weather and illumination conditions and span a range of different scene structures. The university campus is under significant renovation and there are many active constructions areas. The configuration of trees, roads boundaries and buildings have dramatically changed in certain areas over the data collection period. These datasets present a considerable variety of scenarios that we use to evaluate the robustness of semantic segmentation models.

\subsubsection{Local modifications to the semantic model}
Semantic segmentation models are normally trained and validated on publicly available datasets in order to demonstrate their performance. It is well understood however that the model performance often does not translate well into different environments, particularly with changes to the sensor configuration. The process of transfer learning is used to address this problem \cite{itsc_work} by incorporating additional labelled samples from the new environment. For our situation, the model was fine-tuned using a locally collected and labelled semantic segmentation dataset (USYD Dataset) which includes 12 categories: `Sky', `Building', `Column Pole', `Road', `Undrivable Road', `Vegetation', `Sign Symbol', `Fence', `Vehicle', `Pedestrian', `Rider' and `Void'. 
In Fig.~\ref{fig:trajectory}, the green trajectory indicates the route used to collect local images which were subsequently manually annotated using the labelme toolbox~\cite{labelme}. In this USYD Dataset, there are around 200 images recorded by a front-facing PointGrey camera ($56^\circ$ FOV) in mostly cloudy days without strong shadows or over-exposures.

\subsection{Road Surface Extraction Using Lidar}
This section describes the process used to classify the road surface from a lidar point cloud which is subsequently projected onto the image. The outputs of this process are images that are automatically labelled with the semantic `Road' class, which is one of the most important classes for path planning, localisation, driving decision making, etc. in autonomous vehicle operations.

\subsubsection{3D lidar motion correction and filtering}

The lidar provides range and bearing information relative to the frame of the vehicle. As the laser scans mechanically, and the vehicle is moving, the global frame of the sensor changes continuously within each scan which introduces significant distortion. It is important to compensate for the translation and rotation of the vehicle for each lidar point to remove this error~\cite{Himmelsbach2008}. 

The strategy for correcting the motion distortion in each point cloud packet is originally from \cite{supersensor}.
The lidar points are first translated into the vehicle footprint coordinate frame using the rigid transform $T^L_{veh}$. The corrected lidar points can then be evaluated by:
\begin{equation} \label{eq1}
\widetilde{X}_{L,i}=T_{Ego}^{-\Delta _i/\Delta _0}\cdot T^L_{veh} \cdot \overline{X}_{L,i}
\end{equation} where $\overline{X}_{L,i}$ is a raw lidar point, $T_{Ego}$ is the ego-motion transform, and the $\widetilde{X}_{L,i}$ is the final corrected point.

The lidar data collected in urban environment can be contaminated by noise sources including dirt, leaves or other light objects blown by the wind. A statistical filter is applied to remove outliers such as separated points in the air, and other measurement errors. A median filter is also applied to replace the point coordinates with the median of neighboring points.

\subsubsection{Road points extraction}

The road surface is extracted using 6 lidar rings (shown in Fig.~\ref{fig:beam}) that cover a reasonable distance in front of the car. The algorithm presented does not use the lidar return intensity as it is significantly affected by the reflectance of different objects. We consider instead a smoothness factor, evaluated by the change in angle between neighboring points, to separate road and non-road areas. Since the trajectory of the vehicle was free of collisions, it is guaranteed that the space immediately in front of the vehicle is free of obstacles.

To evaluate the smoothness factor, it is necessary to first find the intersections of the projected future vehicle trajectory with the lidar rings. When driving straight, the intersection point is directly in front of the vehicle. When the vehicle is turning, the trajectory can be modelled as a circle of radius: 
\begin{equation}
R=\frac{H}{tan(\alpha)}-\frac{W}{2}\end{equation}
where $\alpha$ is the yaw angle, $W$ and $H$ are vehicle's width and height. The radius $R$ also meets 
\begin{equation}
R^2 = (y^2 - R) + x^2 
\end{equation}
as shown in Fig.~\ref{fig:circle}. Each lidar ring radius $r_n$ can be represented in Fig. \ref{fig:beam} as: 
\begin{equation}
r_n =  r_n^{s} + c
\end{equation}
\begin{equation}
(r_n)^2 = y^2 + (x^2 - d)
\end{equation}
Therefore, the starting point becomes the intersection of the trajectory circle and each lidar ring. 

Since the distance between two circle centers is $D=\sqrt[]{R^2+d^2}$, the triangle area $\partial$ formed by two circle centers and the intersection point $q$ can be calculated as  
\begin{equation}
\partial=\frac{1}{4} \sqrt[]{(q+r_n)(D-R+r_n)(q-r_n)(-D+R+r_n)}
\end{equation}
Then the $Y$ coordinate of the intersection point is:
\begin{equation}
y_n=-\frac{d}{2}-\frac{d(R^2 - r_c^2)}{2D^2}+\frac{R}{D^2}\partial
\end{equation}

After defining the intersection point which is known to be on-road, we start iterating outwards measuring the angle between neighbouring points. If the angle for a given pair of points is larger than the set threshold, the evaluation will be interrupted and this point is considered to be the boundary between the road and non-road area.

\section{Validation Pipeline Implementation}
\label{sec:experiment}

This section describes the implementation details of the validation pipeline including the training of the semantic segmentation model, classifying the lidar `Road' class, and using these measures to validate the semantic model. 

\subsection{Semantic Segmentation Model Training}

The ENet~\cite{enet} architecture for semantic segmentation was used in this paper because of its lightweight and good real-time performance. The model was first trained on the Cityscapes Dataset~\cite{cityscapes}, and then fine-tuned by adding additional training samples from our USYD Dataset using a GeForce GTX 1080 Ti graphics card. Several data augmentation algorithms including flipping, cropping, adding noise, blurring and gamma correction were also applied to artificially expand the dataset and improve the performance in a different domain~\cite{itsc_work, augmentation}.

The input image can have an arbitrary resolution, while the output segmentation result has a fixed resolution of $640 \times 360$. To balance the unequally distributed classes in the dataset, median frequency balancing~\cite{median_balance} was applied to reweigh each class in the cross-entropy loss function. 
The learning rate was set to be $5e-5$ at the beginning and decayed by $0.1$ when the validation error stopped reducing for $100$ epochs.

\begin{figure}[t!]
\centering
\begin{subfigure}[]{0.45\columnwidth}
\centering
	\includegraphics[width=\columnwidth]{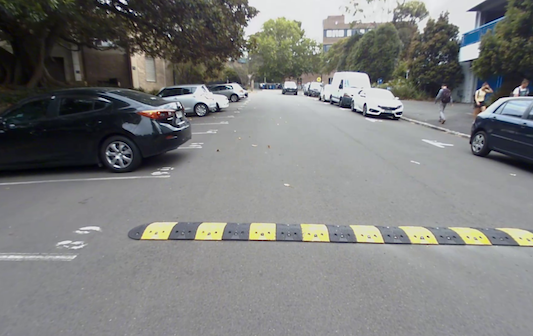}
    \caption{\small Original image}
    \label{sub_a}
    \end{subfigure}
\begin{subfigure}[]{0.45\columnwidth}
\centering
	\includegraphics[width=\columnwidth]{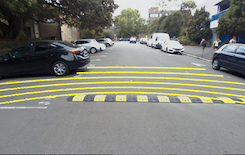}
    \caption{\small One lidar scan}
    \label{sub_b}
    \end{subfigure}  
\caption{\small Lidar points projected to image. \small 
(a) the original image extracted from ROS message (b) 6 lidar beams for road surface extraction. Yellow color indicates lidar points projected onto the image. }
\end{figure}

\subsection{Lidar-based Road Label Generation}
\label{sec:lidar results}

\subsubsection{Sparse lidar point cloud accumulation}
The Velodyne VLP-16 used in the datasets have lidar points across the road area that are extremely sparse compared to the image pixels. The closest lidar points are around two meters away from the vehicle, so when the points are projected into the image frame there is a large space directly in front of the vehicle which has no corresponding lidar information (as shown in Fig.~\ref{sub_b}). Upsampling the point cloud reduces the accuracy of the projection into the image frame, and leads to poor classification performance. To improve the coverage of the lidar information projected to the image, we accumulate multiple lidar scans from before and after the `reference' ($ref$ coordinate) time.

The relative vehicle position and orientation information is obtained from a combination of wheel encoders and a VN-100 IMU containing 3-axis gyros, accelerometers and magnetometers. These odometry measurements are measured relative to the starting position ($odom$ coordinate) of the vehicle. The lidar measurements, lidar point $L_t^{current}(x_t^{current},y_t^{current},z_t^{current})$, are relative to the vehicle frame which can be converted into the $odom$ frame by knowing the $current$ position of the vehicle.

The vehicle pose is represented by a 3 dimensional position vector and a 4 dimensional quaternion vector derived from the odometry information. We denote the position as $P_t^{odom}$ and the quaternion as $q_t$ for each timestamp. The points are concatenated using the process shown below:  
\begin{equation}
P_{t_{\_relative}}^{ref} = q_{t_{\_ref}} \times (P_{t_{\_current}}^{odom} - P_{t_{\_ref}}^{odom}) \times q_{t_{\_ref}}^*
\label{equ1}
\end{equation}
\begin{equation}
q_{t_{\_relative}} = q_{t_{\_current}} \times q_{t_{\_ref}}^*
\label{equ2}
\end{equation}
\begin{equation}
L_{t_{i}}^{ref}(x_{t_{i}}^{ref}, y_{t_{i}}^{ref},z_{t_{i}}^{ref}) = q_{t_{\_relative}} \times L_{t_{i}}^{current} + P_{t_{\_relative}}^{ref}
\label{equ3}
\end{equation}
\begin{equation}
L_{accumulated} = \sum_{1}^{n} L_{t_{i}}^{ref}
\label{equ4}
\end{equation}

Given a reference position $P_{t_{\_ref}}^{odom}$ and reference quaternion $q_{t_{\_ref}}$, we can calculate the Euclidean distance between each past/future position $P_{t_{\_current}}^{odom}$ relative to the current `reference' position. Equation~\ref{equ1} converts this Euclidean distance to the $ref$ coordinate from the $odom$ frame by rotating the vector using the quaternion. Equation \ref{equ2} and \ref{equ3} show the conversion between lidar points at every past/future position to be relative to $ref$ coordinate. The final step (Equation~\ref{equ4}) is to concatenate the lidar points from $n$ lidar messages.

\begin{figure}[t]
\centering{
\includegraphics[width=.85\columnwidth]{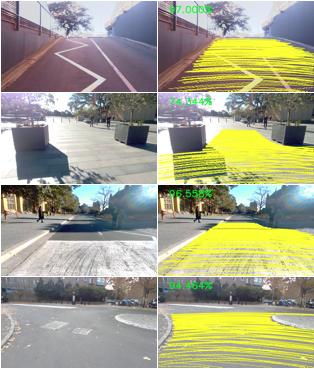}}
\caption{\small Accumulated lidar points (yellow color) projected onto corresponding images. 5 future lidar messages and 20 previous lidar messages were used to generate an accumulated point cloud. The percentages shown in green are the semantic segmentation validation results which will be discussed later.}
\label{fig:lidar_results}
\end{figure}

\begin{figure}[t]
\centering
\begin{subfigure}{.95\columnwidth}
  \centering
  \includegraphics[width=.99\linewidth]{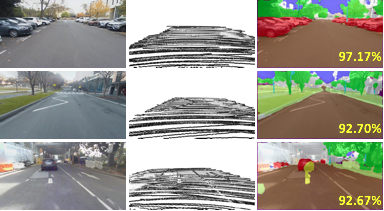}
  \caption{Cloudy weather conditions}
  \label{fig:lidar_val_a}
\end{subfigure}
\begin{subfigure}{.95\columnwidth}
  \centering
  \includegraphics[width=.99\linewidth]{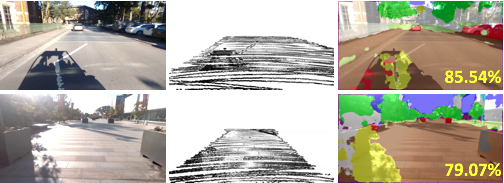}
  \caption{Harsh shadow conditions}
  \label{fig:lidar_val_b}
\end{subfigure}
\begin{subfigure}{.95\columnwidth}
  \centering
  \includegraphics[width=.99\linewidth]{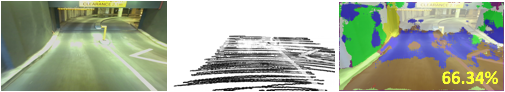}
  \caption{Lighting change at underground car park}
  \label{fig:lidar_val_c}
\end{subfigure}
\begin{subfigure}{.95\columnwidth}
  \centering
  \includegraphics[width=.99\linewidth]{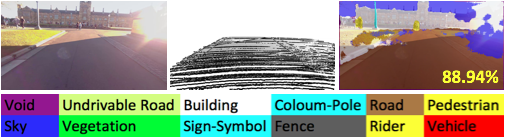}
  \caption{Direct sunlight into camera}
  \label{fig:lidar_val_d}
\end{subfigure}
\caption{Semantic segmentation `Road' class accuracy validated by Lidar. The first column contains the original images extracted from ROS messages. The middle column represents the ground truth labels from lidar extracted road. The third column is the semantic segmentation output with the accuracy of the `Road' class detection overlaid. The percentage value shows how many pixels were correctly classified as `Road' class compared to the lidar ground truth.}
\label{fig:lidar_val}
\end{figure}

To generate the concatenated point cloud, we use the 20 lidar messages prior and 5 lidar messages after the reference timestamp. This covers approximately $\frac{2}{3}$ of the road in the corresponding image when the vehicle is moving. Using the previous lidar messages ensures that the bottom part of image (directly in front of the vehicle) contains samples of the road surface. Incorporating lidar messages taken after the reference time allows samples of the road from further distances to be projected onto the image.

Fig.~\ref{fig:lidar_results} shows the road surface detected using the lidar projected into the reference image as yellow points. Each individual lidar beam was segmented to determine the boundary between the road and pavement or other obstacles by considering when the angle between neighboring points was larger than a certain threshold. The pixels from the reference image corresponding to the lidar road points are then converted into a one-channel labelled image and used as the ground truth to compare the output of the semantic segmentation model described in the next section.

\subsubsection{Lidar points to image projection}

After concatenating several lidar point clouds, the accumulated points are projected onto the corresponding image frame using homogeneous transformations and camera projections:

\begin{equation}
\begin{bmatrix}
x' \\
y' \\
z'
\end{bmatrix}
= 
C \times T_{lidar}^{camera}
\begin{bmatrix}
X_{l}\\
Y_{l}\\
Z_{l}\\
1
\end{bmatrix} 
=
\begin{bmatrix}
f_xX_{c} +o_xZ_{c} \\
f_yY_{c} +o_yZ_{c} \\
Z_{c}
\end{bmatrix}
\end{equation}
where $C = \begin{bmatrix}
f_x & 0& o_x & 0 \\
0&f_y&o_y&0\\
0&0&1&0
\end{bmatrix}$ is the camera intrinsic matrix and $T_{lidar}^{camera}$ is the transformation matrix to transform lidar coordinate to camera coordinate. Then the corresponding pixel $(u, v)$ can be found by:
\begin{equation}
\begin{bmatrix}
u \\
v \\
1
\end{bmatrix}
= 
\begin{bmatrix}
f_xX_{c}/Z_{c} +o_x \\
f_yY_{c}/Z_{c} +o_y \\
1
\end{bmatrix}
\end{equation}

The area covered by these points are considered to be the ground truth for road surface as determined by the lidar. 

\begin{figure*}[!t]
\centering
\includegraphics[width=.95\textwidth]{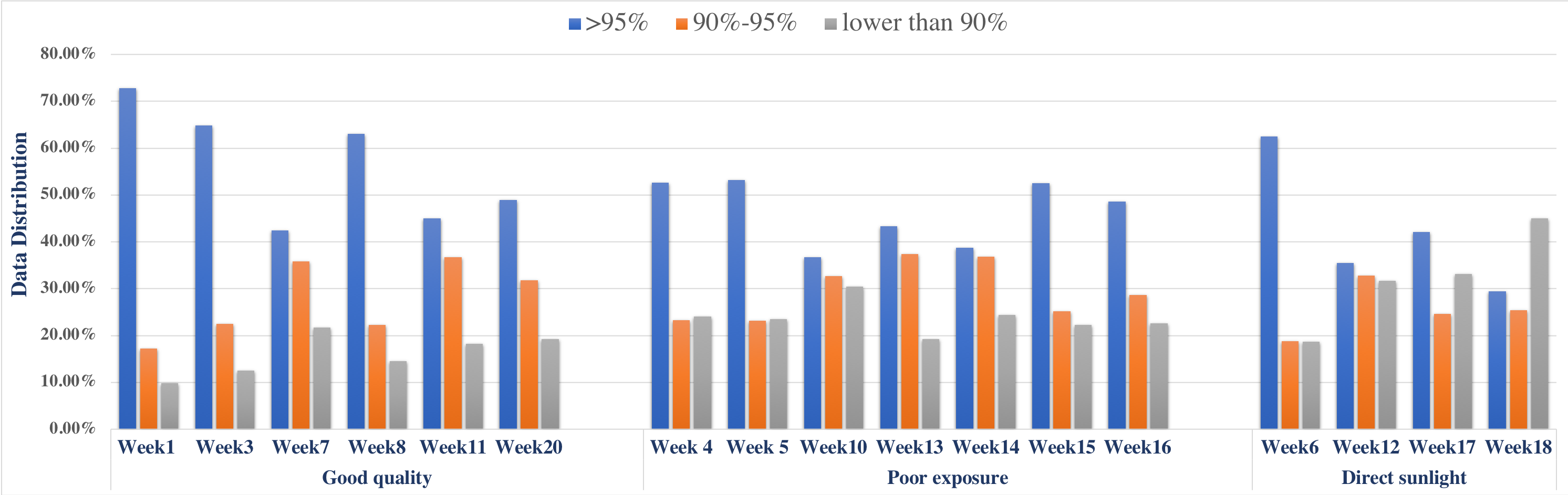}
\caption{Distribution of validation results related to various environmental and camera conditions. The image data is visually examined and manually grouped into `Good quality', `Poor exposure' and `Direct sunlight' based on a joint consideration of weather, illumination, sensor performance and sun locations.}
\label{fig:camera condition}
\end{figure*}

\begin{figure}[!t]
\centering
\includegraphics[width=.8\columnwidth]{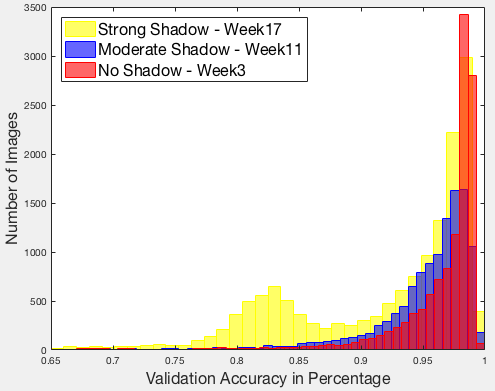}
\caption{Shadow influence on semantic segmentation performance. }
\label{fig:distribution}
\end{figure}

\subsection{Semantic Segmentation Validation}
\label{sec:semantic results}
The semantic segmentation model is validated following the pipeline described in Fig.~\ref{fig:pipeline}. The first process in the pipeline involves extracting the ROS messages to obtain the lidar point cloud, camera image, GPS location, odometry and ROS transform information for a given reference position. The next stage is the accumulation of the lidar point clouds from the set of messages before and after the reference time. This requires the projection of each point into the odometry frame of the reference position. Once transformed, the points can then be projected into the corresponding image and used as the ground truth for road surface. In parallel with this stage, the camera image at the reference position is published to the semantic segmentation node to complete the segmentation process. The validation process takes the ground truth image and compares with the output of the semantic segmentation model. The validation percentage is calculated by comparing the area of the lidar classified road surface label with the semantic segmentation results:
\begin{equation}
\frac{N(semantic\_road\_pixel | lidar\_road\_area)}{N(lidar\_road\_points)}
\end{equation}
The final step is to associate the validation percentage with the location of the sample given by a satellite-based global position.
During this process, all of the global locations and validation percentages are stored into a PostgreSQL~\cite{postgre} database and plotted against a satellite image using QGIS~\cite{qgis} (results are shown in Sec.~\ref{sec:gps_associ}).

\section{Experimental Results}
\label{sec:Experimental Results}

In this section, we show the results of the automated semantic segmentation validation process in a range of different conditions and in different locations. We also show how this process can be used to compare the results of different network architectures.

\subsection{Validating Performance across Datasets}

The proposed validation pipeline is used to determine how many pixels have been correctly classified as `Road' class compared with the lidar-based ground truth labels. As the lidar points are sparser than image pixels, the percentage correct metric only considers the parts of the image corresponding to projected lidar points.

\begin{figure*}[t]
\centering
\begin{subfigure}{.45\textwidth}
  \centering
  \includegraphics[width=\linewidth]{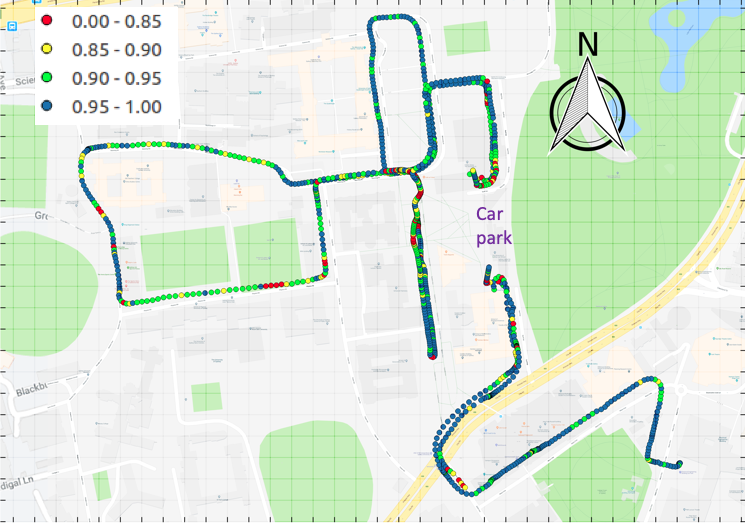}
  \caption{\small Week Number: 10. Date: 2018-05-15 Time: 1pm-2pm. Weather: Sunny. Shadow: Strong.}
  \label{fig:week10}
\end{subfigure}
~
\begin{subfigure}{.45\textwidth}
  \centering
  \includegraphics[width=\linewidth]{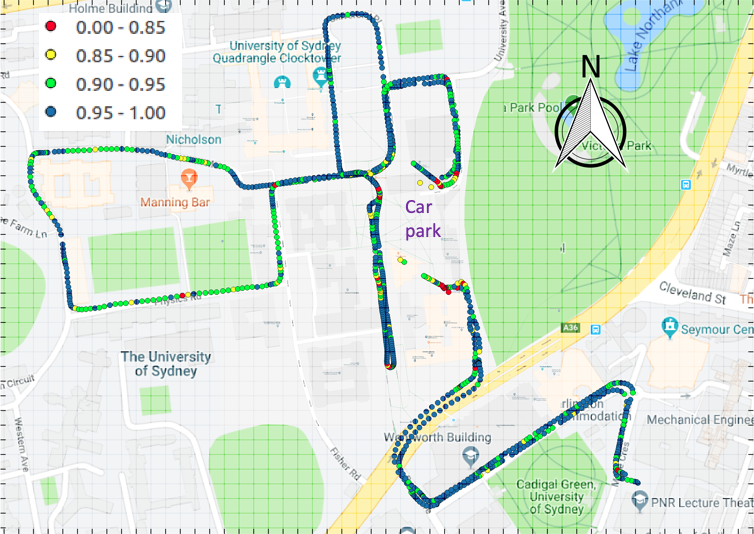}
  \caption{\small Week Number: 11. Date: 2018-05-28 Time: 11am-12pm. Weather: Cloudy. Shadow: Moderate.}
  \label{fig:week11}
\end{subfigure}
\caption{\small `Road' class semantic segmentation validated by lidar extracted road surface and associated with GPS locations. Visualized in Qgis. GPS information is lost in the underground car park. }
\label{fig:GPS_associated}
\end{figure*}

\begin{figure*}[t]
\centering
\begin{subfigure}{.45\textwidth}
  \centering
  \includegraphics[width=\linewidth]{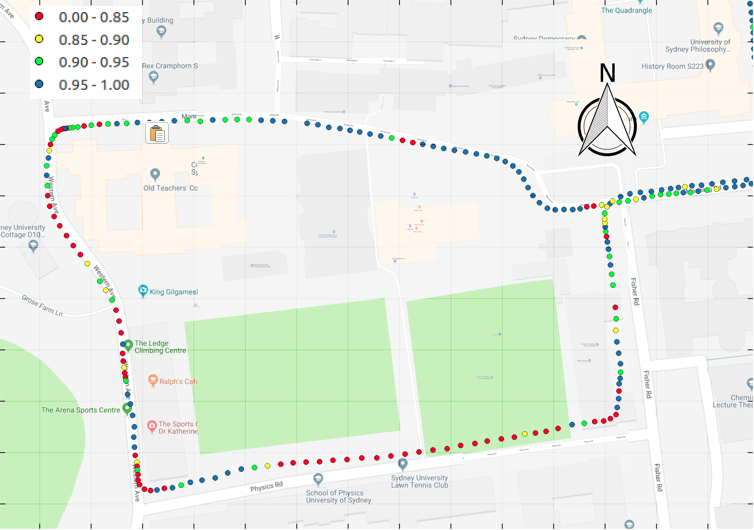}
  \caption{\small Week16 Dataset sub-area. ENet model fine-tuned on USYD Dataset with four randomly selected data augmentation methods~\cite{itsc_work}.}
  \label{fig:week16_itsc}
\end{subfigure}
~
\begin{subfigure}{.45\textwidth}
  \centering
  \includegraphics[width=\linewidth]{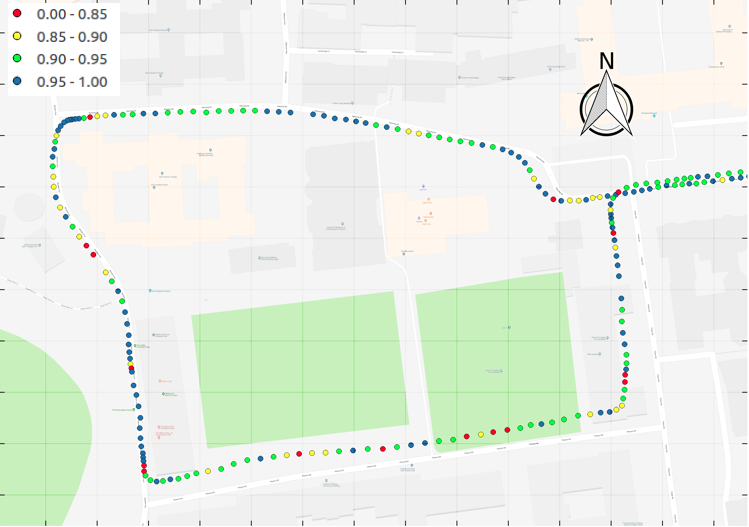}
  \caption{\small Week16 Dataset sub-area. Existing ENet model~\cite{itsc_work} fine-tuned with gamma correction augmentation was used to improve robustness~\cite{augmentation}.}
  \label{fig:week16}
\end{subfigure}
\caption{\small Model improvement validation. A small area was taken from Week16 Dataset to demonstrate the improvement after applying the  gamma corrected augmentation training samples.}
\label{fig:improvement}
\end{figure*}

\subsubsection{Qualitative validation results across local environmental conditions}

As the datasets were regularly collected at different times of the year, images in these datasets cover a wide variety of environmental conditions as illustrated in Fig.~\ref{fig:lidar_val}. 

In Fig.~\ref{fig:lidar_val_a}, the selected testing images have roughly similar illumination condition as the training images in USYD Dataset. The high cloud cover restricted the contrast caused by shadows and resulted in favourable illumination conditions. 
Subsequent datasets were collected with a change in the camera hardware, and with a wide variety of weather/illumination conditions.
The main difference with the new datasets was the high contrast between shadows and direct sunlight on bright days. The camera configuration and the contrast caused problems for the semantic segmentation model and resulted in a higher percentage of false positives particularly for the `Road' class (Fig.~\ref{fig:lidar_val_b}).
These new datasets operated outside the area of the training data, and included a drive through an underground car park. Transitioning between outdoors and the underground car park resulted in dramatic changes in light levels, and the camera requires some time to automatically adjust to the new conditions. The internal car park lighting is also different to sunlight, and the reflections from the artificial lighting on the ground resulted in a higher percentage of false positives (Fig.~\ref{fig:lidar_val_c}). 
Another issue that was identified in several datasets occurred when the sun was low in the sky, a problem that is more prominent in winter due to the lower trajectory of the sun through the sky. In this situation, there were a number of locations where the sun directly shone into the camera lens causing large areas of over-exposure in the image (Fig.~\ref{fig:lidar_val_d}). The semantic segmentation performed poorly when the sun was orientated in this way.

By visually inspecting a wide range of the qualitative results and grouping them into several bands, it appears that images with higher than 95\% accuracy generally represent acceptable segmentation results, even with some minor noise at the object boundaries. A percentage range between 90\% and 95\% generally indicate some noise outside the object boundaries and a percentage range between 85\% and 90\% starts indicates a more perceptible error. When the accuracy is lower than 85\%, there is generally some significant classification failure in the image which requires further improvement to the model. 

\subsubsection{Quantitative analysis of validation results} 

As we have a diverse range of environmental and camera conditions, we visually inspected the representative image quality from each dataset and manually grouped them into three categories based on a combined consideration of weather, illumination, sun location and camera performance: 
\begin{itemize}
    \item `Good quality' usually indicates cloudy days with no challenging illumination conditions and objects in the image generally have clear boundaries;
    \item `Poor exposure' commonly happens in very sunny days with harsh illumination conditions and strong reflections into the camera which results in the existence of over-exposures in most images;
    \item `Direct sunlight' occurs when the sun location is too low and there are direct sunlight and lens flares in the image as shown in Fig.~\ref{fig:lidar_val_d}.
\end{itemize}

In Fig.~\ref{fig:camera condition}, we summarise the validation results based on these three categories. When the camera is performing well with good lighting conditions, the semantic segmentation has better performance in general. In these datasets, there are a higher percentage of images with classification performance greater than $90\%$. With the remaining dataset categories, i.e. when there are many poorly exposed images or with the lens under direct sunlight, the number of points with less than 90\% accuracy increases dramatically, with the exception of Week 6 which can be considered as an outlier. 

\subsubsection{Detailed analysis of the accuracy distribution}
Shadows have been universally recognized as a strong factor affecting the performance of vision-based algorithms. In this section, we select three representative datasets: one with strong shadows, one with moderate shadows and one with almost no shadows. We analyse these datasets in more detail to demonstrate the influence of shadows on semantic segmentation performance. 

Fig.~\ref{fig:distribution} shows the detailed distribution of the classification performance for each of the three datasets using a histogram of the validation accuracy. The red bars show the distribution of classification percentage from a dataset with almost no shadows on the ground due to heavy cloud cover. The performance distribution is heavily skewed towards $100\%$ with many samples measured as greater than $98\%$ accuracy. The dataset with moderate shadows (blue bars), shows the distribution is shifted slightly lower towards $96\%$ accuracy. The dataset with strong shadows (yellow bars) is seen to be a bimodal distribution which the top mode is above $95\%$, and a smaller mode peaking around $84\%$. 
The bimodal nature of the strong shadow dataset is caused by specific locations in the dataset where the consequences of direct sunlight or strong shadows influence the camera. This happens primarily in particular areas with partial shade (i.e. under trees), which also indicates that there is a spatial component to the classification performance. The spatial context will be further explored in the next subsection.

\subsection{Validating Performance in a Spatial Context}
\label{sec:gps_associ}

To evaluate the performance of a semantic segmentation model in a geospatial context, we associate the validation percentage metric obtained in the previous sections with a corresponding global position obtained from a GNSS sensor.

\subsubsection{Overall performance}
Fig.~\ref{fig:GPS_associated} shows a plot of the validation percentage metric where the percentage range is indicated by different colours. The blue dots correspond to a validation accuracy over 95\%, the green dots are measurements between 90\% and 95\%, the yellow dots are between 85\% and 90\%, and the red dots are lower than 85\%. 
Semantic segmentation models (and classification models in general) are considered to work better in areas where there is labelled training data in comparison to unseen areas. 
Our validation pipeline quantitatively shows this result, with areas incorporating labelled image data in the model (the overlapping green and red trajectories in Fig.~\ref{fig:trajectory}) containing a higher number of blue dots ($>$95\% accuracy) compared to areas outside the training set with comparatively worse performance. 

In addition, the geolocated validation performance metrics are useful to determine areas in which the semantic segmentation performs poorly. In Fig.~\ref{fig:GPS_associated}, although position information is lost in the underground car park, it is still clear that the semantic segmentation has reduced performance (indicated by red and yellow points) at the entrance and exit of car park due to the significant illumination changes. 
The areas that are identified as having poor semantic segmentation performance require further analysis, and cannot be used for autonomous driving until the performance is improved.

\subsubsection{Validating network improvement based on associated GPS information}
In our previous work~\cite{augmentation} we explored the use of data augmentation techniques, specifically gamma correction, to improve the model robustness for images with strong shadows. For the results of this previous work however, we were only able to benchmark the improvement against a limited hand-labelled dataset containing high contrast shadows to demonstrate qualitative results of the augmentation performance.

The negative performance of the semantic model caused by shadows is not consistent across the entire dataset. With the validation pipeline proposed in this paper, we are now able to identify the issues with the segmentation model in specific areas. The model is then re-trained using data augmentation techniques to address the issues specific to the areas in question. Finally, the proposed pipeline is able to quantitatively demonstrate the network improvement under the these conditions. 

Fig.~\ref{fig:improvement} illustrates this process using a specific area of the dataset to compare the differences in the semantic model performance with additional augmented training samples.
Fig.~\ref{fig:week16_itsc} shows the spatial performance of the ENet model trained with randomly selected augmentation algorithms including flipping, cropping, adding noise and blurring~\cite{itsc_work}. It is clear that there are a significant number of red points in certain areas indicating that the performance of the network is poor. Fig.~\ref{fig:week16} shows the change in performance after adding gamma corrected data augmentation~\cite{augmentation}. There is a significant improvement after the gamma corrected training samples are added, with a reduction in the number of low-accuracy points (red points).

\subsection{Comparison of Different Networks}

Deep neural networks are generally benchmarked against public datasets to compare their performance. These datasets cover only a limited range of locations, scene participants, lighting and weather conditions which may not be sufficient to examine the ability of a network to generalise across a range of real-world scenarios. By using the proposed validation pipeline, we compared ENet~\cite{enet} and Bonnet~\cite{bonnet} and demonstrate their performance under a wide variety of conditions. 

The standard method for evaluating two networks is to compare performance against the same dataset and benchmark them using metrics such as mean accuracy, computation performance, etc. The original Cityscapes Dataset~\cite{cityscapes} has more than 30 classes. A number of these classes are not relevant to the autonomous vehicle application, so we remapped this set of classes into 12 categories and trained ENet and Bonnet using this simplified dataset.

\begin{table}[h]
\definecolor{Gray}{gray}{0.9}
\normalsize
\begin{center}
\caption{Standard comparison between networks on Cityscapes dataset. `mCA' is mean class accuracy and `mIoU' is mean Intersection over Union, `Road' is road class accuracy, and `Speed' is the model inference runtime tested on NVIDIA DRIVE PX2 with the image resolution of $640 \times 360$. }
\label{tab:standard}
\begin{tabular}{| l | l | l | l | l |}    
\hline
\rowcolor{Gray}
     & mCA & mIoU & `Road' & Speed \\ \hline
    \textbf{ENet} & 82.93\% & 66.78\% & 95.47\% & 27 fps \\ \hline
    \textbf{Bonnet} & 91.66\% & 68.37\%  &95.76\% & 26 fps \\
    \hline
\end{tabular}
\end{center}
\end{table}

\begin{table*}
\definecolor{Gray}{gray}{0.9}
\normalsize
\caption{Bonnet and ENet comparison on four weeks datasets. The percentage shown in the last column is the averaged validation percentage across all data in each dataset.}
\label{tab:multiple}
\begin{center}
\begin{tabular}{|c|c|c|c|c|c|}
\hline
\rowcolor{Gray}
Week No.&	Date&Starting Time&	Image Condition&	Shadow&	Average Percent \\ \hline

12&2018-06-04&	09:19&Direct sunlight&	Strong& \shortstack{\textbf{93.81\% ENet} \\ 93.05\% Bonnet}
 \\ \hline
 
14&	2018-06-18&	11:01&Good quality&	No & \shortstack{\textbf{94.23\%
 ENet} \\ 93.72\% Bonnet}    \\ \hline

17&2018-07-10&14:14&Poor exposure&	Strong	& \shortstack{91.22\% ENet \\\textbf{92.70\% Bonnet}} \\ \hline

18&2018-07-17 &14:14 &Poor exposure&	Strong	& \shortstack{90.12\% ENet \\ \textbf{90.40\% Bonnet}} \\ \hline

\end{tabular}
\end{center}
\end{table*}

Table \ref{tab:standard} shows the standard comparison results between networks. The results show that the Bonnet model has a higher average class accuracy and mean Intersection over Union measurement, while the `Road' class accuracy and computation speed are similar to the ENet model. Based on this result, the Bonnet model would be selected over ENet given a superior global performance when measured against the public datasets.

To provide a more comprehensive analysis of the performance of each model under different environmental conditions, we fine-tuned both the Bonnet and ENet models using our locally labeled USYD Dataset. We then compared the performance of the two networks using our validation pipeline using datasets covering multiple weeks. 

We selected four datasets with varying conditions to demonstrate the comparison between the Bonnet and ENet models, shown in Table~\ref{tab:multiple}. 
The results show that Bonnet has better performance during bright sunny days with higher image contrast caused by strong shadows in the images. 
The results are reversed however on dataset from days where the sunlight is not so strong, and the images are of a better quality with fewer high contrast areas. In this case, the ENet model demonstrated a higher average percentage of correct classifications.
This is an important result to show that the `best' network can be dependent on the dataset, and it is not possible to form a complete comparison without taking many varied datasets into consideration.

\section{Conclusion and Future Work}
\label{sec:Conclusions}
Semantic segmentation has been a popular area of research in recent years with the development of deep neural networks and high performance hardware for computation. Validating the reliability and robustness of a trained model under different environmental conditions has not been widely addressed due to the complexity of generating ground truth labels which generally requires labour-intensive hand labelling. In this paper, we have addressed the validation problem by proposing a pipeline which incorporated an additional sensor (lidar) to examine the semantic segmentation performance of a model in a variety of real-world scenarios. By examining the geometric properties of neighbouring lidar points, we were able to distinguish the boundaries of the road near to the vehicle, and automatically generate a large amount of ground truth road labels. By comparing the `Road' class from the output of the semantic segmentation process against the lidar generated ground truth, we have shown that it is possible to obtain a measure of the classification accuracy in order to validate the model. 

We have collected a weekly dataset around our local area across different times of the year. These datasets were used to perform a comprehensive analysis of the performance of a trained segmentation network. In addition, the validation accuracy of a model was compared against datasets with different weather conditions, lighting conditions, geospatial context, etc. This analysis was used to show the variation in performance given a range of conditions and to compare/visualise the performance in different locations.

The validation pipeline was also used to compare the difference in performance between two different semantic models. The standard approach to comparing different deep neural networks is to measure the performance against publicly available datasets which provide only a limited range of conditions. This can potentially lead to a suboptimal outcome when selecting the best model for use in a different real-world application. 
We have demonstrated that the selection of the best model can depend on the operating conditions, the relative accuracy of two models can vary depending on the dataset.
In addition, we used the validation pipeline to visualise the improvements resulting from applying data augmentation to the training dataset. The spatial context of the performance metrics were used to identify areas with poor performance, and after applying the appropriate data augmentation it was shown that the performance in these areas was improved.

The importance of system validation in an engineering system cannot be understated. Before semantic segmentation models can be used in real-world applications, it must be proven that the model is capable of acceptable performance under different environmental conditions.
We have demonstrated that it is possible to identify areas where additional training is required before the model is suitable for autonomous operation. 
The proposed validation pipeline not only enables large scale system validation without requiring hand labelled data, but has the potential to provide a metric of performance in real time while the vehicle is in operation.
This pipeline can be further extended to cover additional sensor modalities, and to automatically label other classes to create additional robustness to semantic segmentation in an autonomous driving context.

\section*{Acknowledgment}
This work has been funded by the Australian Centre for Field Robotics (ACFR), Australian Research Council Discovery Grant DP160104081, and the University of Michigan/Ford Motor Company Contract `Next Generation Vehicles'.

\bibliographystyle{IEEEtran}
\bibliography{Biblio}

%
\begin{IEEEbiography}[{\includegraphics[width=1in,height=1.25in,clip,keepaspectratio]{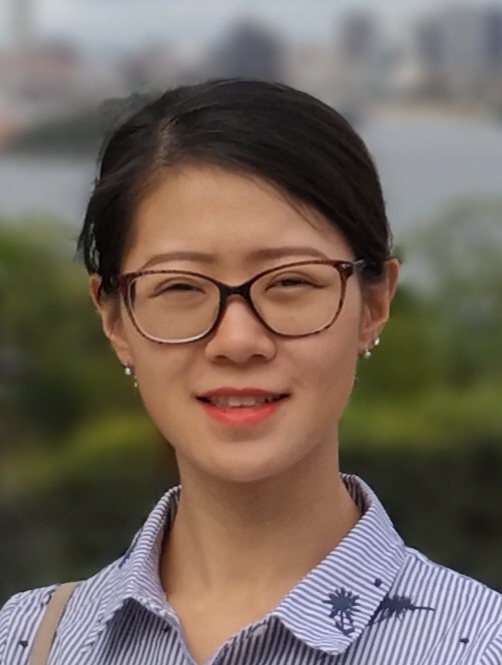}}]{Wei Zhou} received double B.S. degrees in Electronic and Communication Systems from Beijing Institute of Technology and Australian National University in 2013 , and the M.E. degree in Telecommunications from Australian National University in 2014. She is currently working towards her Ph.D. degree at the University of Sydney, Sydney, Australia. Her research is focused on real-time semantic segmentation for autonomous vehicles. 
\end{IEEEbiography}

\begin{IEEEbiography}[{\includegraphics[width=1in,height=1.25in,clip,keepaspectratio]{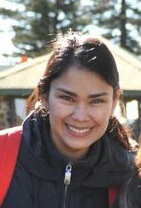}}]{Julie Stephany Berrio} received the B.S. degree in Mechatronics Engineering in 2009 from Universidad Autonoma de Occidente, Cali, Colombia, and the M.E. degree in 2012 from the Universidad del Valle, Cali, Colombia. She is currently working towards the Ph.D. degree at the University of Sydney, Sydney, Australia. Her research interest includes semantic mapping, long-term map maintenance, and point cloud processing.
\end{IEEEbiography}

\begin{IEEEbiography}[{\includegraphics[width=1in,height=1.25in,clip,keepaspectratio]{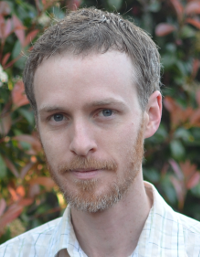}}]{Stewart Worrall} received the Ph.D. from the University of Sydney, Australia, in 2009. He is currently a Research Fellow with the Australian Centre for Field Robotics, University of Sydney. His research is focused on the study and application of technology for vehicle automation and improving safety.
\end{IEEEbiography}

\begin{IEEEbiography}[{\includegraphics[width=1in,height=1.25in,clip,keepaspectratio]{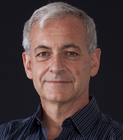}}]{Eduardo Nebot} received the BSc. degree in electrical engineering from the Universidad Nacional del Sur, Argentina, M.Sc. and Ph.D. degrees from Colorado State University, Colorado, USA. He is currently a Professor at the University of Sydney, Sydney, Australia, and the Director of the Australian Centre for Field Robotics. His main research interests are in field robotics automation. The major impact of his fundamental research is in autonomous systems, navigation, and safety.
\end{IEEEbiography}




\end{document}